\documentclass[11pt,a4paper]{article}
\usepackage[hyperref]{acl2020}
\usepackage{times}
\usepackage{latexsym}

\usepackage{microtype}
\usepackage{graphicx}
\usepackage{subcaption}
\usepackage{url}

\aclfinalcopy 

\title{Sentence Embeddings using Supervised Contrastive Learning}

\author{Danqi Liao \\
  Princeton University \\
  \texttt{dl33@princeton.edu} \\}

\date{}

\begin{document}
\maketitle
\begin{abstract}
Sentence embeddings encode sentences in fixed dense vectors and have played an important role in various NLP tasks and systems.  Methods for building sentence embeddings include unsupervised learning such as Quick-Thoughts\cite{quickthoughts} and supervised learning such as InferSent\cite{infersent}. With the success of pretrained NLP models, recent research\cite{sbert} shows that fine-tuning pretrained BERT \cite{bert} on SNLI\cite{snli:emnlp2015} and Multi-NLI\cite{MULTINLI18-1101} data creates state-of-the-art sentence embeddings, outperforming previous sentence embeddings methods on various evaluation benchmarks.

In this paper, we propose a new method to build sentence embeddings by doing supervised contrastive learning. Specifically our method fine-tunes pretrained BERT on SNLI data, incorporating both supervised cross-entropy loss and supervised contrastive loss. Compared with baseline where fine-tuning is only done with supervised cross-entropy loss similar to current state-of-the-art method SBERT\cite{sbert}, our supervised contrastive method improves 2.8\% in average on Semantic Textual Similarity (STS) benchmarks\cite{sts-cer-etal-2017-semeval} and 1.05\% in average on various sentence transfer tasks.
\end{abstract}

\section{Introduction}
BERT-based pretrained language models\cite{bert} have achieved state-of-the-art results on various sentence classification and regression tasks. Traditional method for using BERT to solve sentence and sentence pair tasks feeds the input sentence or sentence pair to the BERT model and then passes the output of BERT to a simple classification or regression function to output the final prediction. Recent research \cite{sbert} proposes sentence-BERT models that can build semantically meaningful independent sentence embeddings for single sentences. \cite{sbert} fine-tunes modified BERT on SNLI and MultiNLI data to achieve state-of-the-art sentence embeddings.

In this work, we present sentence embeddings using supervised contrastive learning\cite{khosla2020supervised}. Specifically our method builds on top of the SBERT network proposed by \cite{sbert}. We fine-tunes the networks on SNLI data using both cross-entropy loss and supervised contrastive loss. Supervised contrastive loss is computed by comparing positive and negative sentences for each anchor sentence directly in sentence embedding space. We discuss the model architectures and supervised contrastive learning objectives in details in Section "Model". 

Compared with the baseline similar to SBERT where fine-tuning is only done using cross entropy loss, our best models using supervised contrastive learning improve 2.8\% in average on Sentence Textual Similarity (STS) benchmarks and 1.05\% in average on various sentence transfer tasks. 

\section{Related Work}
\subsection{Sentence Embeddings}

Early methods for constructing sentence embeddings involve directly taking the average or weighted pooling of word embeddings of tokens in input sentence \cite{arora2b29624234f9441cab6dd9c918e86ab7}. Moving beyond simple pooling method,  Skip-Thought\cite{kiros2015skipthought} is proposed to build sentence embeddings using unsupervised objective. Very much similar to Skip-Gram method for building word embeddings\cite{mikolov2013distributed}, Skip-Thought method builds sentence embeddings by predicting surrounding sentences around a target sentence. The model used in Skip-Thought is RNN-based encoder-decoder networks. An improvement on Skip-Thought in terms of training time is Quick-Thought\cite{quickthoughts}. Quick-Thought reframes the context sentence generation problem to sentence classification problem: it replaces the decoder with a classifier that picks next sentence among a batch of candidates.

More recent research attempts on building sentence embeddings by using supervised learning. InferSent\cite{infersent} trains a bi-LSTM sentence encoder and a classifer on top of it using labelled SNLI data. The sentence embeddings given by InferSent encoder is able to outperform the ones constructed by previous unsupervised sentence embedding methods. 

There are also attempts in combining both unsupervised and supervised training objective. Universal Sentence Encoder trains transformer based network\cite{vaswani2017attention}, using both supervised and unsupervised objectives, on various datasets and various sentence tasks.

The current state of the art sentence embeddings method is proposed by \cite{sbert}. They fine-tune pretrained BERT \cite{bert} on SNLI\cite{snli:emnlp2015} and Multi-NLI\cite{MULTINLI18-1101} data and their experiments show that their SBERT model is able to outperform previous sentence embeddings methods on various evaluation benchmarks including Sentence Textual Similarity (STS) benchmarks and a variety of sentence transfer tasks.

\subsection{Supervised Contrastive Learning}

Contrastive learning has recently gained momentum in self-supervised representation learning in computer vision tasks\cite{chen2020simple}. Going beyond unsupervised contrastive learning, supervised contrastive learning\cite{supercontrast} is proposed to efficiently leverage the label information in labelled datasets. Points of the same class are pulled together while points of different classes are pushed apart in embedding space. Image recognition model trained using supervised contrastive method is able to outperform the state-of-art model on top-1 accuracy on ImageNet Dataset.

\cite{gunel2020nlpsupercontrasts} adapts supervised contrastive learning objective on fine-tuning pretrained language models on various NLP tasks and shows that models fine-tuned with supervised contrastive learning are more robust to noisy train data and have significant gains compared to the baseline in few-shot learning settings.

\begin{figure}[t]
    \centering
   \includegraphics[scale=0.35]{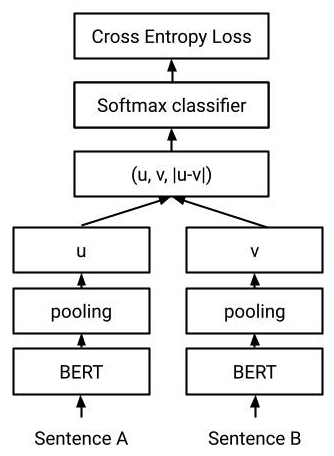}
    \caption{Model structure for computing cross-entropy loss on SNLI data. The two BERT networks have tied weights.}\label{fig:ce1}
\end{figure}

\begin{figure}[t]
    \centering
   \includegraphics[scale=0.28]{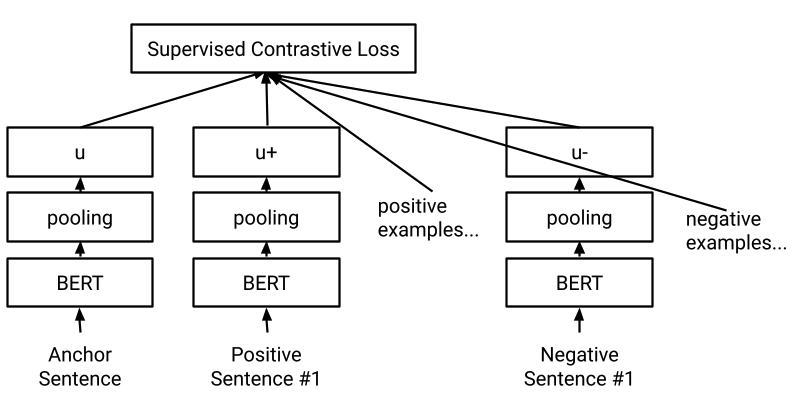}
    \caption{Model structure for computing supervised contrastive loss on SNLI data. All BERT networks have tied weights.}\label{fig:scl1}
\end{figure}
\begin{table*}[t]
   \begin{tabular}{|c|c|}
 \hline
Anchor &\texttt{\small A brown dog is laying on its back on the grass with a ball in its mouth.}\\
 \hline
  Positive    & \texttt{A dog is laying on is back outside.} \\
  Positive  & \texttt{A dog is on the grass with his toy.} \\
  Positive  & \texttt{A brown dog has a ball in its mouth.}\\
 \hline
    Negative  &\texttt{A black dog is using the restroom in the park.} \\
    Negative  &\texttt{a brown dog is runnig with a stick.}\\
    Negative  &\texttt{there is a dog eating food off the table.}\\
  \hline
\end{tabular}
\caption{Positive and negative sentences for example anchor sentence \texttt{A brown dog is laying on its back on the grass with a ball in its mouth.} from SNLI train dataset.}
    \label{tab:my_label}
\end{table*}
\section{Model}
Our model is based on the SBERT network proposed by \cite{sbert}. We compute both the cross-entropy loss and supervised contrastive loss. The model structures of cross-entropy loss and supervised contrastive loss are different.  

\paragraph{Cross Entropy Loss} The model structure for computing standard cross-entropy loss is almost identical to the one used in SBERT (Figure~\ref{fig:ce1}). We feed the input sentence pair individually to the BERT model and then pool the output token embeddings except the first \texttt{[CLS]} token. After pooling, we get the sentence embeddings for the input sentences and we feed the concatenation of the two sentence embeddings and their absolute difference to a two-layer softmax classifier. The cross-entropy loss is computed on the softmax classification scores. The formula for computing cross-entropy loss given a batch of input sentence pairs $x$ is:
\begin{equation}
      \mathcal{L}_{CE} =-\frac{1}{N}\sum_{i=1}^{N}\sum_{c=1}^{C}y_{i,c} \cdot \log \hat{y}_{i,c}
\end{equation}
here $N$ denotes the number of data in a batch, $C$ is the number of total classes, $y_{i,c}$ is the ground truth label for sentence pair $x_i$ belonging to class $c$. $\hat{y}_{i,c}$ is the predicted probability of  sentence pair $x_i$ belonging to class $c$.

\paragraph{Supervised Contrastive Loss} The model structure for computing supervised contrastive loss is illustrated in Figure~\ref{fig:scl1}. We treat each input sentence in a batch as an anchor sentence and we find the positive and negative sentences of that anchor sentence. We discuss the criterion on selecting and constructing positive and negative sentences in next paragraph. We feed the anchor sentence and its positive and negative sentences individually to the BERT model. We pool the BERT output except \texttt{[CLS]} token to get sentence embeddings for each sentence. With the sentence embeddings of each sentence, we compute the supervised contrastive loss for single anchor sentence in a batch as the following:
\begin{equation}
      \mathcal{L}_{SCL} = -\frac{1}{N_{i+}} \sum_{j=1}^{N_{i+}} \log \frac{e^{{\Phi(x_i}) \cdot \Phi(x_j) / \tau}}{\sum_{k=1}^{N_{i+,i-}} e^{{\Phi(x_{i}) \cdot \Phi(x_{k})/ \tau}}}, 
\end{equation}
where $\Phi(x_i)$ denotes the sentence embeddings for input sentence $x_i$ produced by the model; $N_{i+}$ is the number of positive sentences for specific anchor sentence $x_i$ and $N_{i+,i-}$is the number of positive and negative sentences for anchor sentence $x_i$; $\tau$ is the temperature hyperparameter for softmax function that measures the separation of classes.

Intuitively, supervised contrastive loss tries to pull positive sentences toward anchor sentence while push away negative sentences from the cluster of positive sentences directly in embedding space. 

The loss of our models is computed as a linear interpolation between cross-entropy and the supervised contrastive learning loss,
\begin{equation}\label{new_ob}
 \mathcal{L} = (1- \lambda) \cdot \mathcal{L}_{CE} + \lambda \mathcal{L}_{SCL}
\end{equation}
$\lambda$ is the weight for supervised contrastive learning loss term.
\begin{table*}[t]
\centering
\begin{tabular}{ |c|c|c|c|c|c|c| }
 \hline
\textbf{Model} & \textbf{STS12} & \textbf{STS13} & \textbf{STS14} & \textbf{STS15} &\textbf{STS16} & \textbf{Avg.}\\
 \hline
  Avg. GloVe embeddings & 44.09 & 43.02 &47.43 &50.08 &40.3 &44.98  \\
  SBERT baseline & \textbf{66.06} & 63.41 &67.03 &74.46 & 67.11 &67.61 \\
  3p3n-Augment & 65.85 & 62.76 &66.2 &73.1 &64.97 & 66.58\\
  allpalln-Augment & 65.68&65.04&66.81&72.68&63.28 & 66.70\\
 \hline
    3p3n-lambda0.5-SCL & 64.1& 63.81& 66.13& 74.11& 67.69 &67.17 \\
     3palln-lambda0.5-SCL &65.05&68.34&69.08&76.09&70.22 & 69.76\\
     allpalln-lambda0.5-SCL &64.89 &69.42 &69.29 &75.69&70.98 &70.05\\
    3all3n-lambda1.0-SCL& 62.71&61.87&65.65&72.49&70.12&66.57 \\
     3palln-lambda1.0-SCL &63.44&68.71&69.57&75.57&71.5 & 69.76\\
     allpalln-lambda1.0-SCL & 64.17&67.14&69.27&75.73&70.61 & 69.38\\
     allpalln-lambda0.1-SCL & 65.72&67.08&67.99&75.68&69.45 &69.18\\
     allpalln-lambda0.3-SCL &65.21&\textbf{69.48}&\textbf{69.73}&\textbf{76.14}&\textbf{71.63} &\textbf{70.44}\\
     
  \hline
\end{tabular}
\caption{Evaluation results on STS tasks. Weighted Spearman correlations are reported for each STS task. Average score across all STS tasks are list at the last column. Model Avg. GloVe embeddings takes the average GloVe word embeddings for all tokens in a sentence. SBERT baseline is the baseline model fine-tuned using only cross-entropy objective. 3p3n-Augment, allpalln-Augment are other baseline models and are trained on augmented anchor, positive and negative sentences using pure cross-entropy loss. 3p3n means at most 3 positive sentences and at most 3 negative sentences are selected for each anchor sentence. 3palln means at most 3 positive sentences and all negative candidates are selected for anchor sentence. allpalln means all positive and negative sentence candidates are selected. Models named with lambda1.0 are trained only on supervised contrastive loss.}\label{tab:accents}

\end{table*}
\paragraph{Anchor, Positive and Negative Sentence} To compute supervised contrastive loss, we need to construct the anchor, positive and negative sentences. For each sentence pair from SNLI dataset, there is a premise and a hypothesis. The relationship between the premise and hypothesis is either entailment, neutral or contradiction. For a batch of input sentence pairs from SNLI dataset, we treat each premise as one single anchor sentence. Positive sentences of the anchor sentence are the hypotheses that have entailment relation with the anchor sentence. Negative sentences are the hypotheses that have either neutral or contradiction relation with the anchor sentence, or hypotheses that are paired with other premises in the batch. 

\section{Experimentation}
\subsection{Data}
We train all of our models including the baseline on 550,152 data from SNLI train dataset\cite{snli:emnlp2015}. SNLI is a natural language inference dataset that has English sentence pairs manually annotated with the labels entailment, neutral and contradiction. In SNLI train dataset, each unique premise appears at least three times with different hypotheses. 
\subsection{Implementation \& Experiments Setup}
We use pretrained 8-layer uncased BERT with hidden size 512\cite{midbert} as our BERT model in the networks. The pretrained BERT model is implemented by Google and made available to use on HuggingFace platform\cite{wolf2019huggingfaces}. We use Adam optimizer with learning rate $2e^-5$ and a linear optimizer scheduler with 10\% of the train data as warm-up step. All models are trained using batch size 64 and only trained with 1 epoch. The pooling method used in all of our models including baseline is mean pooling.  It takes around 30-36 minutes to train a single model on a single Tesla V100 GPU. Our model source code is available at \url{https://github.com/Danqi7/584-final}.

\subsection{Evaluation}
Following previous work and what's been done in SBERT\cite{sbert}, we evaluate our models on Semantic Textual Similarity (STS) benchmarks and various sentence transfer tasks. It takes around 28 minutes to evaluate one single model. All of model evaluation are done using SentEval\cite{conneau2018senteval}. SentEval is a public available evaluation toolkit for benchmarking universal sentence representations.
\paragraph{Semantic Textual Similarity (STS)} We evaluate our models on Semantic Textual Similarity (STS) benchmarks in an unsupervised setting. Specifically the evaluation is done on STS tasks from 2012, 2013, 2014, 2015 and 2016\cite{agirre-etal-2012-semeval, agirre-etal-2013-sem, agirre-etal-2014-semeval, agirre-etal-2015-semeval, agirre-etal-2016-semeval}. STS tasks contain pairs of sentences collected from various sources and each sentence pair is labelled with score between 0 and 5 based on how semantically related the pair is. We evaluate how the cosine similarity distance between two sentences embeddings correlate with a human-labeled similarity score through Pearson and Spearman correlations. Since research\cite{reimers-etal-2016-task} shows that Pearson correlation is badly suited for STS evaluation, we report weighted Spearman’s rank correlation for each STS task.

\begin{table*}[t]
\centering
\begin{tabular}{ |c|c|c|c|c|c|c|c|c|c| }
 \hline
\textbf{\small Model} & \textbf{\small MR} & \textbf{\small CR} & \textbf{\small SUBJ} & \textbf{\small MPQA} &\textbf{\small SST2}&\textbf{\small SST5} &\textbf{\small TREC} &\textbf{\small MRPC} & \textbf{\small Avg.}\\
 \hline
  \small Avg. GloVe embeddings & 51.14 &78.78&99.6&\textbf{87.59} & \textbf{79.68} &43.8 &82.8&70.78&74.27\\
  \small SBERT baseline & 66.1&80.34&\textbf{99.62}&86.92 &79.08&38.96&83.8&69.86&75.56\\
  \small 3p3n-Augment &62.28 & \textbf{81.06}&99.56&86.86 & 77.65& \textbf{44.71}&83.8&69.62&75.69\\
  \small allpalln-Augment &63.24 &79.95 &99.6 &86.88&76.22&40.86&88.2&71.77 &75.84 \\
 \hline
\small 3p3n-lambda0.5-SCL & 64.76&80.45&99.6&84.78&75.45&42.31&\textbf{89.2}&72.12&76.08\\
\small 3palln-lambda0.5-SCL&64.57&80.29&99.6&87.38&76.99&41.67&87.6&72.99& 76.37\\

\small allpalln-lambda0.5-SCL & \textbf{67.43}&80&99.6&87.12 &74.85& 40.72&87.4&\textbf{73.1}&76.28\\

\small 3all3n-lambda1.0-SCL&60.57&79.47&99.6&86.53 &74.63&41.04&87.8&71.48&75.14\\
\small 3palln-lambda1.0-SCL &67.33&80.27&99.6&86.84 &76.06& 40.18&85.8&72.17&76.03\\
\small allpalln-lambda1.0-SCL &56.57&79.87&99.6&86.87 & 76.39&41.45&87.4&71.59&74.97\\
\small allpalln-lambda0.1-SCL & 66.19&80.21&99.6&86.94&77.1&42.26& 87.8&73.04& \textbf{76.64}\\
\small allpalln-lambda0.3-SCL& 64.67&79.68&99.6&86.99&76.77&42.53&86.8&72.36 & 76.16\\
  \hline
\end{tabular}
\caption{Evaluation results on SentEval transfer tasks. The average scores of all tasks are listed at the last column.}\label{tab:r2}
\end{table*}

\paragraph{SentEval Transfer Tasks} We also evaluate our models on various sentence classification tasks using toolkit SentEval in a supervised setting. Sentence embeddings constructed by the embedding models are used as features to fit a simple classifier on various sentence classification datasets. For most of the classification tasks, 10-fold validation is used for computing the test accuracy. We evaluate our models on the following eight SentEval transfer tasks:
\begin{itemize}
    \item MR: Sentiment analysis for movie reviews\cite{pang-lee-2005-seeing}.
    \item CR: Sentiment prediction for product reviews\cite{Hu2004MiningAS}. 
    \item SUBJ: Subjectivity analysis for movie reviews and plot summaries \cite{10.3115/1218955.1218990}.
    \item MPQA: Opinion polarity analysis for new articles from a wide range of news sources\cite{Wiebe2005AnnotatingEO}.
    \item SST Binary:  Stanford Sentiment Treebank with binary labels\cite{socher-etal-2013-recursive}.
    \item SST Fine-Grained: Stanford Sentiment Treebank with 5 labels\cite{socher-etal-2013-recursive}.
    \item TREC: Question classification prediction\cite{10.3115/1072228.1072378}.
    \item MRPC: Semantic equivalence analysis for sentence pairs\cite{10.3115/1220355.1220406}.
\end{itemize}

\subsection{Results}
We train various models with different hyperparameters. Specifically, we investigate three hyperparameters including $N_{+}$(the number of positive sentences for anchor sentence), $N_{-}$ (the number of negative sentences for anchor sentence), $\lambda$ (the weight for supervised contrastive loss term in training objective) and $\tau$ (the the temperature term for computing softmax). Note that $N_{+}$ does not mean exact $N_{+}$ positive sentences and $N_{-}$ does not mean exact $N_{-}$ negative sentences because in the context of SNLI dataset, some anchor sentences have less than $N_{+}$ positive sentences. The number of positive sentences $N_{+}$ is upper-bounded by the number of positive candidates in a batch. The number of negative sentences $N_{-}$ is upper-bounded by the number of negative candidates in a batch. 

All models listed in this section have a fixed temperature value 1.0. The best model on STS tasks is our supervised contrastive learning model with all positive and negative candidates, lambda 0.3 and temperature 1.0 (allpalln-lambda0.3-SCL); Model allpalln-lambda0.3-SCL outperforms SBERT baseline with 2.8\% improvement in average. The best model on SentEval transfer tasks is supervised contrastive learning model with all positive and negative candidates, lambda 0.1 and temperature 1.0 (allpalln-lambda0.1-SCL). Model allpalln-lambda0.1-SCL outperforms SBERT baseline with 1.05\% improvement in average. Ablation studies on these three hyperparameters are discussed in the next section. 
\begin{table*}[h]
\centering
\begin{tabular}{ |c|c|c|c|c|c|c|c|c|c| }
 \hline
\textbf{\small $\lambda$} & \textbf{\small MR} & \textbf{\small CR} & \textbf{\small SUBJ} & \textbf{\small MPQA} &\textbf{\small SST2}&\textbf{\small SST5} &\textbf{\small TREC} &\textbf{\small MRPC} & \textbf{\small Avg.}\\
 \hline
  \small 0.1 & 66.19&80.21&99.6&86.94&77.1&42.26& \textbf{87.8}&73.04& \textbf{76.64}\\
  \small 0.3 & 64.67&79.68&99.6&86.99&76.77&\textbf{42.53}&86.8&72.36 & 76.16\\
  \small 0.5 & \textbf{67.43}&80&99.6&\textbf{87.12} &74.85& 40.72&87.4&\textbf{73.1}&76.28\\
  
  \small 0.7 & 60.7& \textbf{81.09} &99.6&87.05& \textbf{76.88} &40.5&86.6&72.35&75.60 \\
\small 0.9 & 61.91&79.2&99.58&86.95&75.12&41.72&87.2&70.38 &75.26\\
  \hline
\end{tabular}
\caption{Ablation experiment results for different $\lambda$ on SentEval transfer tasks .All models here have a fixed $N_+$, $N_-$ and share the same temperature value 1.0.}\label{tab:ablation2}
\end{table*}
\begin{table*}[h]
\centering
\begin{tabular}{ |c|c|c|c|c|c|c|c|c|c| }
 \hline
\textbf{\small $\tau$} & \textbf{\small MR} & \textbf{\small CR} & \textbf{\small SUBJ} & \textbf{\small MPQA} &\textbf{\small SST2}&\textbf{\small SST5} &\textbf{\small TREC} &\textbf{\small MRPC} & \textbf{\small Avg.}\\
 \hline
  \small 0.1 & \textbf{67.62}&78.31&99.6&87.22&73.59&41.27&86.4&71.42&75.68\\
  \small 0.3 & 60.76&79.65&99.6&\textbf{87.44}&76.66&\textbf{43.21}&86.8&72.81&75.87\\
  \small 0.5 & 64.76&80.77&99.6&86.77&75.67&41.99&\textbf{88.4}&70.26&76.03\\
  
  \small 0.7 &58&80.78&99.6&86.43&75.29&39.86&87.8&\textbf{73.85}&75.20 \\
  
\small 0.9 & 63.62&\textbf{81.35}&99.58&86.97&\textbf{78.14}&41.31&83.4&71.88&75.78\\
\small 1.0 &67.43&80&99.6&87.12 &74.85& 40.72&87.4&73.&\textbf{76.28}\\

  \hline
\end{tabular}
\caption{Ablation experiment results for different temperature $\tau$ on SentEval transfer tasks .All models here have a fixed $N_+$, $N_-$ and share the same $\lambda$ value 0.5.}\label{tab:temp2}
\end{table*}
\begin{table}[t]
\centering
\resizebox{\columnwidth}{!}{
\begin{tabular}{ |c|c|c|c|c|c|c| }
 \hline
\textbf{$\lambda$} & \textbf{STS12} & \textbf{STS13} & \textbf{STS14} & \textbf{STS15} &\textbf{STS16} & \textbf{Avg.}\\
 \hline
  0.1 & \textbf{65.72}&67.08&67.99&75.68&69.45 &69.18  \\
  0.3 &65.21&\textbf{69.48}&\textbf{69.73}&\textbf{76.14}&\textbf{71.63} &\textbf{70.44} \\
  0.5 & 64.89 &69.42 &69.29 &75.69&70.98 &70.05\\
  0.7 & 64.51&68.42&69.36&75.4&70.78 &69.69\\
    0.9 & 64.47&68.39&69.32&75.76&70.11 &69.61 \\
  \hline
\end{tabular}}
\caption{Ablation experiment results for different $\lambda$ on STS tasks .All models here have a fixed $N_+$, $N_-$ and share the same temperature value 1.0. }\label{tab:ablation}

\end{table}


\paragraph{Semantic Textual Similarity (STS)}We present our model evaluation results on STS tasks in Table~\ref{tab:accents}. Sentence embeddings generated using supervised contrastive learning outperforms the baselines on almost every STS task except STS12. The best model is allpalln-lambda0.3-SCL with a 2.8\% improvement on SBERT baseline in average. Increasing the number of positive and negative sentences for each anchor sentence increases the STS evaluation scores. 

All augment models trained using pure cross-entropy loss perform worse than the SBERT baseline, suggesting the gain using supervised contrastive loss does not simply come from more data. Moreover, 3p3n-lambda0.5-SCL and 3p3n-Augment have the same amount of augmented positive and negative sentences, but 3p3n-lambda0.5-SCL outperforms 3p3n-Augment. allpalln-lambda0.5-SCL outperforms allpalln-Augment model with a 3.35\% improvement in average. The performance improvement from using supervised contrastive learning with the same amount of data suggests that our supervised contrastive objective is more efficient than cross-entropy loss in building sentence embeddings.

\paragraph{SentEval transfer tasks} We present our model evaluation results on eight SentEval Transfer Tasks in Table~\ref{tab:r2}. Majority of models trained using supervised contrastive learning outperform the baseline models in average. The best model is allpalln-lambda0.1-SCL, outperforming SBERT baseline with a 1.05\% improvement in average. Baseline models including SBERT baseline, 3p3n-Augment and Avg. GloVe embeddings have better scores than SCL models on some transfer tasks such as SST2, CR and SST5.

The improvement using contrastive supervised learning evaluated on supervised sentence transfer tasks is smaller than the improvement using contrastive supervised learning evaluated on unsupervised sentence tasks. We hypothesize the smaller improvement may be because in supervised settings, the model can fine-tune the extra classifier on top of the sentence embeddings, therefore the model can learn to perform well on the transfer tasks even with relatively worse sentence embeddings.

\section{Ablation Studies}
In this section, we perform ablation experiments on 3 hyperparameters of our sentence embedding method in order to better understand their relative importance.

\subsection{Effect of Lambda}
We explore the effect of $\lambda$, the weight on supervised contrastive loss term, on the produced sentence embeddings (Table ~\ref{tab:ablation} and Table~\ref{tab:ablation2}). For STS tasks, the best model has $\lambda$ value 0.3. Increasing $\lambda$ from 0.5 to 0.9 decreases the quality of generated sentence embeddings, suggesting a combination of cross-entropy and supervised contrastive loss generate better sentence  embeddings than individual losses. For SentEval transfer tasks, the best model  has $\lambda$ value 0.1. Increasing $\lambda$ tends to decrease the model performance. 
\begin{table}[t]
\centering
\resizebox{\columnwidth}{!}{
\begin{tabular}{ |c|c|c|c|c|c|c| }
 \hline
\textbf{$\tau$} & \textbf{STS12} & \textbf{STS13} & \textbf{STS14} & \textbf{STS15} &\textbf{STS16} & \textbf{Avg.}\\
 \hline
  0.1& 59.71&56.6&60.28&67.82&63.66 &61.61\\
 0.3 &63.12&62.84&65.09&72.74&66.79&66.12 \\
0.5 & 64.09&65.75&67.6&74.79&68.96 &68.24\\
  0.7& 64.56&67.3&69&75.46&70.61 &69.39\\
  0.9 & 64.57&69.2&69.2&75.62&70.3&69.78\\
    1.0 &\textbf{64.89} &\textbf{69.42} &\textbf{69.29} &\textbf{75.69}&\textbf{70.98} &\textbf{70.05} \\
  \hline
\end{tabular}}
\caption{Ablation experiment results for different temperature $\tau$ on STS tasks. All models here have a fixed $N_+$, $N_-$ and share the same $\lambda$ value 0.5 }\label{tab:temp}

\end{table}
\subsection{Effect of Temperature}
We explore the effect of temperature $\tau$, the temperature term for
computing softmax, on sentence embedding models (Table ~\ref{tab:temp} and Table~\ref{tab:temp2}). Smaller temperature value creates harder negatives. For STS tasks, the best model has $\tau$ value 1.0. Increasing $\tau$ from 0.1 to 1.0 increases the quality of generated sentence embeddings. For SentEval transfer tasks, the best model has $\tau$ value 1.0. Increasing $\tau$ from 0.1 to 0.5 and 0.7 to 1.0 tend to increase the model performance but there is a drop going from 0.5 o 0.7.

\subsection{Effect of $N_+$, $N_-$}
We show the effect of $N_+$, $N_-$, the number of positive and negative sentences for each anchor sentence, in above section Results. The results are summarized in Table~\ref{tab:accents} and Table~\ref{tab:r2}. For both STS tasks and SentEval transfer tasks, the best model has $N_+$ the number of all positive candidates and $N_-$ the number of all negative candidates. Models achieve best evaluation result when all positive and negative candidates are used for each anchor sentence. For both STS tasks and SentEval transfer tasks, increasing $N_+$ and $N_-$ tends to increase evaluation scores in general, with an exception that going from 3 $N_+$, all $N_-$ to all $N_+$, all $N_-$ with lambda value 1.0 (using only supervised contrastive loss term), the performance score drops around 1.0\% in average on SentEval transfer tasks.

\section{Conclusion}
We propose a new method to build sentence embeddings by fine-tuning pretrained BERT on SNLI data using supervised contrastive learning. We empirically show that sentence embeddings generated using supervised contrastive learning consistently outperform embeddings generated by the baselines using only cross-entropy loss. We also demonstrate that simply fine-tuning on augmented data using cross-entropy loss does not generate better sentence embeddings than the baselines. Our best models using supervised contrastive learning outperform the SBERT baseline with 2.8\% improvement in average on unsupervised STS tasks and 1.05\% improvement in average on various supervised sentence transfer tasks.

In addition, building sentence embeddings using supervised contrastive learning is computationally efficient. It only take 36 minutes in total to train our best model. The training time is just 16 extra minutes compared to training the SBERT baseline.

\bibliography{acl2020}
\bibliographystyle{acl_natbib}

\end{document}